\documentclass[runningheads]{llncs}

\usepackage{multirow}
\usepackage{algorithm}
\usepackage{amsmath}
\usepackage[graphicx]{realboxes}
\usepackage{nccmath}
\usepackage{algpseudocode}
\usepackage{amssymb}
\usepackage{amsmath,graphicx,amssymb,braket,xcolor,subfigure,upgreek,bbold}
\usepackage{multirow}

\addtocontents{toc}{~\vspace{-3\baselineskip}}
\usepackage{tikz}

\usepackage{indentfirst}
\usepackage{array, boldline, rotating}
\usepackage{etoolbox}
\usepackage{rotating}
\usepackage{amsmath,siunitx}
\newcommand{\nrange}[4][]{%
  #2=%
  \ifblank{#1}{
    #3, \dots ,#4%
  }{%
    #3, #1, \dots,#4%
  }%
}

\begin{document}

\title{ICU Mortality Prediction using Long Short-Term Memory Networks}
\titlerunning{ICU MP using LSTM Networks}
%

\author{Manel Mili\inst{1,2}\orcidID{0000-0003-3892-8579} \and
Asma Kerkeni\inst{1,2}\orcidID{0000-0002-3134-3802}  \and
Asma Ben Abdallah\inst{1,2}\orcidID{0000-0001-7821-7734}  \and
Mohamed Hedi Bedoui\inst{1}\orcidID{0000-0003-4846-1722} }
\authorrunning{Mili et al.}
%
\institute{Faculty of Medicine, University of Monastir, Tunisia\\
\and
Higher Institute of Computer Sciences and Mathematics,  University of Monastir, Tunisia\\
\email{
manel.mili@isimm.u-monastir.tn}
\email{asma.kerkeni@isimm.u-monastir.tn}
\email{asma.benabdallah@isimm.u-monastir.tn}
\email{medhedi.bedoui@fmm.rnu.tn}}

\maketitle              
\begin{abstract}
Extensive bedside monitoring in Intensive Care Units (ICUs) has resulted in complex temporal data regarding patient physiology, which presents an upscale context for clinical data analysis. In the other hand, identifying the time-series patterns within these data may provide a high aptitude to predict clinical events. Hence, we investigate, during this work, the implementation of an automatic data-driven system, which analyzes large amounts of multivariate temporal data derived from Electronic Health Records (EHRs), and extracts high-level information so as to predict in-hospital mortality and Length of Stay (LOS) early. Practically, we investigate the applicability of LSTM network by reducing the time-frame to 6-hour so as to enhance clinical tasks. The experimental results highlight the efficiency of LSTM model with rigorous multivariate time-series measurements for building real-world prediction engines.

\end{abstract}
\section{Introduction}
An ICU serves patients with severe complications or life-threatening injuries, which involve constant care in order to maintain normal bodily functions. To improve hospital services, it seems important to adequately select patients to be admitted to ICUs early on. In an ICU, the patient is monitored using Electronic Health Record (EHR) systems, entering many medical data a day including physiological measurements. Finding statistic models in these measurements has the potential to provide a high aptitude for more accurate and earlier predictions of future clinical events. This might not only help clinicians make more effective medical decisions but also facilitate an economical allocation of hospital resources. Naturally, mortality prediction and Length of Stay (LOS), are mainly performed with an interest in the prediction of possible outcomes, which are the death or survival of the patient, and for how long a patient may remain in the intensive units. Nevertheless, most available mortality and LOS prediction systems \cite{le1993new,simpson2016new,Pirracchio2016,awad2019predicting} in the literature were designed for at least 24-hour to provide a real-time or retrospective prediction on patients’ mortality. To enhance prediction for early diagnosis, the main objective of this paper is to develop an end-to-end approach based on deep learning models, within a data mining framework, specifically intended for predicting mortality and LOS, based on multivariate time-series physiological measurements from the first few hours of admission, in particular after the first 6 hours of a patient’s acceptance in the ICU. The rest of the paper is organized as follows: Section \ref{rel} provides a comprehensive literature review on the state-of-the-art works. Section \ref{data} details the process of dataset collection and preparation. Section \ref{three} discuss the proposed model and presents its configuration and implementation tools. To consider the effectiveness of the proposed method, section \ref{exp} deals with experiments. Ultimately, section \ref{con} concludes the paper and highlights the fundamental contributions.

\section{Related Works}\label{rel}
Over the past few decades, substantial researches are undertaken to affect predicting mortality risk and LOS tasks. A number of the more frequently used mortality prediction models in an ICU setting include SAPS-II \cite{le1993new} and SOFA \cite{simpson2016new}. SAPS-II was designed to estimate the probability of mortality, while SOFA was wont to describe organ dysfunction.
Using the primary 24-hour patient physiological measurements, these scores are only designed to form one prediction. As a result, it's unknown how well each system predicts mortality following the primary day of admission. Moreover, it seems intuitively likely that straightforward clinical judgment also will discriminate more effectively as time passes. Existing tools are therefore slow to succeed in useful discriminatory effectiveness and aren't generally felt by clinicians to be useful to help decision-making once they will discriminate.

Adding to severity scores, several authors have converged on management mortality risk, as an example, Pirracchio et al. \cite{Pirracchio2016} aimed to develop a scoring procedure to predict mortality in ICUs supported Super-Learner (SL) model. They have proved that the SL method improved performance. However, the authors evaluated the performance of SL using data recorded within the primary 24-hour. Moreover, Darabi et al. \cite{Darabi2018} developed a model supported Gradient Boosted Tree (GBT) and Convolutional Neural Network (CNN) to estimate the mortality risk of patients admitted to ICUs. Their results prove usability a smaller number of features which will generate satisfactory outcomes for GBT, unlike, CNN that need a wealthy amount of knowledge for training. However, their model was designed within the period of 30-day after admittance.

In addition to mortality risk prediction, few researchers have converged to estimate LOS. Mentioning Gentimis et al. \cite{los} who explored the utilization of Neural Network (NN) for predicting the entire LOS of a patient within the hospital. The predictive model outperforms machine learning models. However, the studied scenarios considered time-frames $> 5$ days, or $\leq 5$ days, to validate the potency of the model. Furthermore, Zebin et al. \cite{length} applied an Auto-Encoder (AE) along side a dense neural network technique attempted at identifying short and long stays for patients. The proposed model improved the performance compared to employing a simplistic dense neural network for the classification task. However, their assessment results were validated using recordings observed after 24 hours of admission.

To conclude, all the above-mentioned works only focused on predicting the risk of mortality and LOS for patients who required intensive care within a minimum of 24-hour of their ICU admission \cite{johnson2012patient,Pirracchio2016,Purushotham2017,Darabi2018,awad2019predicting}. The challenge, therefore, lies within the early hours of a patient’s admission, for instance, the primary 6 and 12 hours. Additionally, not all critically ill patients can enjoy ICU admission. Hence, determining the priority of patients' treatments by the severity of their condition is crucial because the ICU is extremely costly with limited resources. The challenge, therefore, lies in triaging patients consistent with their medical conditions, while estimating their expected time of hospitalization. Adding to the present , most research has centered on the evaluation of the efficiency of their predictive models using univariate time-series data and that they didn't consider the potency of multivariate time-series records for improving the accuracy and therefore the efficiency of time-series modeling \cite{Patrick}.

\section{Dataset}\label{data}
This effort is conducted over the well-known publicly available, large-scale ICU database, the MIMIC-III \cite{Johnson2016}, which presents a single-center electronic database developed by the MIT Lab for Computational Physiology, comprising health data related to $61.532$ ICU admissions of $46.520$ distinct de-identified patients admitted between 2012 and 2020.

\subsection{Feature Engineering}
Every day,  different vital signs measurements are computed and analyzed during intensive stays. In this proceeding, we focused primarily, in hidden patterns within ICU time-series data and investigated the hypothesis that there is much useful knowledge in motifs within these data that can aid to improve prediction clinical tasks. This hypothesis is motivated by observations considered within several studies, for example, in \cite{physionet}, we found that in the event of a lack of oxygen transport, measurements in this time-frame of associated variables increase the risk of death. We therefore explored some temporal variables defined in acuity severity scoring systems and added others since they have proven to possess a powerful effect in predicting mortality and hence LOS \cite{vold2015low}. These variables include "heart rate", "systolic BP", "diastolic BP", "mean BP", "respiratory rate", "oxygen saturation", "glasgow coma score", "blood urea nitrogen", "temperature", "white blood cells", and last not least "bilirubin".

Some of the foremost pertinent measurements could also be obtained using information available within the earliest phase \cite{awad2019predicting}. So, we've extracted features for the primary $6$ hours for every ICU stay. We have also extracted features for the 12 and 24 hours so as to verify the effectiveness of the proposed model in maintaining its accuracy for long periods.
\vspace{-3mm}
\subsection{Feature Preprocessing}
EHRs contain valuable information for estimating mortality risk and discharge time for ICU patients, but substantial missing and imbalanced data present mutual problems for the development and implementation of a prediction model. Hence, the subsequent two issues were identified and handled accordingly.
\vspace{-3mm}
\subsubsection{Missing Data Imputation}
The percent of missing values for certain features is higher than $50$\%.
To manage this problem, data imputation was performed including two strategies: we start by filling them using linear interpolation on each multivariate time-series data. Some observations are still missing after this imputation since there are missing data for certain variables. Hence, we impute missing observations using the Mean as the second strategy.
\vspace{-3mm}
\subsubsection{Imbalanced Data Regulation}
The number of patients who passed away inside the intensive department is relatively small in comparison with the number of patients who survived, yielding an imbalanced dataset. To manage this problem, re-sampling methods were adopted since they are less sensitive to outliers than other techniques like Cost-sensitive classifiers \cite{perry2015imbalanced} and Automatic support vector data description \cite{sadeghi2018automatic}. Two of the most common categories of re-sampling methods are under-sampling and over-sampling strategies. The former remove observations from the training dataset that belong to the dominant class, while the latter duplicate samples that belong to the lesser class, thus increasing its impact within the training process. We have applied the former on the dataset since the latter would make models inflexible in learning during the training process by causing overfitting. As a result, the size of the data was reduced from $33.6$ Mo to $7.76$ Mo, from $66,7$ Mo to $15$ Mo and from $129$ Mo to $29$ Mo, over the 6-hour, 12-hour and 24-hour time-frames, respectively.

\section{Methodology}\label{three}
\begin{figure}[tbh!]
\centering
\includegraphics[width=1\columnwidth]{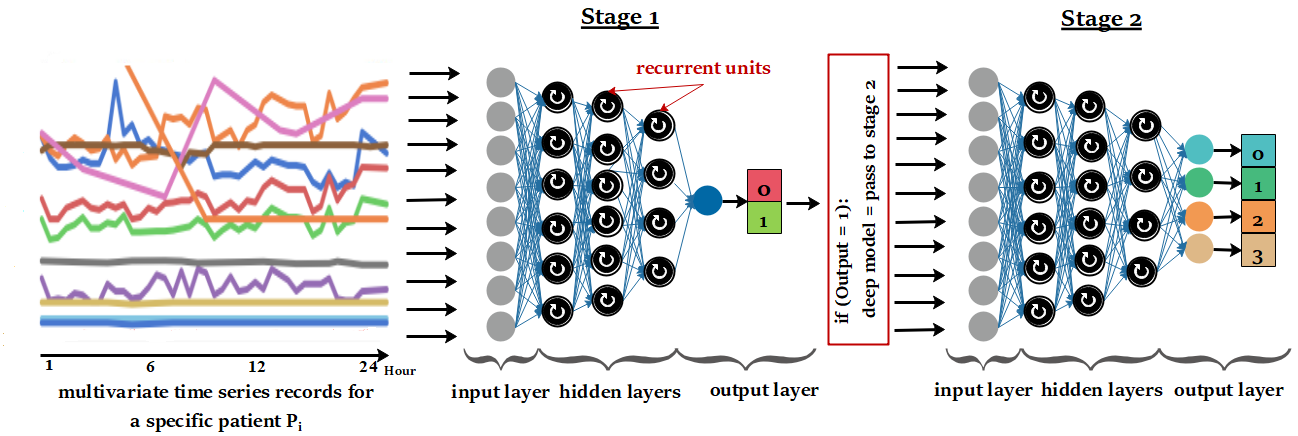}
\caption{A summary structure of two-stage architecture: : within the first stage, a binary classifier is trained to predict mortality. Then, if the mortality is predicted to be positive, the model would further provide an estimation about LOS.}
\label{figg}
\end{figure}

The idea behind time-series prediction is to predict future events supported past values with reference to historical measurements and associated patterns. Turning to the philosophy of the research methodology, we would like to hold relevant information throughout the processing of medical data sequences, as physiological variables begin to decrease or increase over a period of your time, thus making it possible to predict future outcomes associated with patient conditions in care units. To reach these specific goals, a typical two-stage architecture is presented in Fig. \ref{figg}.

The philosophy behind the defined architecture is detailed as follows: we start by interpreting multivariate time-series of $11$ past clinical records for each patient $P_i$:
\begin{equation}  P_i: X_{1,t_{k}}, X_{2,t_{k}}, ..., X_{11,t_{k}}  \end{equation} with $\nrange{k}{1}{n}$ and $n \in$ \{6-hour, 12-hour, 24-hour\}. In the first stage, a binary classifier is trained to predict the risk of mortality. In a mathematical interpretation we identify:
    \begin{align*}
    Class = \begin{cases}
    0, & \text{survivors group}\\
    1, & \text{non-survivors group}.
    \end{cases}
    \end{align*}

\noindent Therefore, we define a knowledge set of two exclusion criteria: we start by filtering by $ 16\leq$ age $\leq 89$ \cite{Purushotham2017}. Then , we exclude ICU stays of but one hour to get rid of obscurity in data due to unusual short stays. After filtering, we observe $49.632$ ICU stays of $36.343$ patients.
While a multi-class classifier is trained at the second stage using a similar vital signs so as to predict LOS for those that are predicted dead in stage $1$. Accordingly, we filter the ICU stays with death time $\leq 0$. As a result, $5.718$ in-hospital mortalities were obtained. We then label each data to at least one of the four classes represented below:\\
 \begin{align*}
Class = \begin{cases}
  0, & \text{if } death\_time\_hours < 6, \\
  1, & \text{if } 6 \leq  death\_time\_hours < 12, \\
  2, & \text{if } 12 \leq death\_time\_hours < 24, \\
  3, & \text{otherwise}.
\end{cases}\end{align*}

The proposed model will predict outcomes values by identifying short-term (6 hours/12 hours) and long-term (24 hours) dependencies. For this purpose, we have employed the LSTM architecture \cite{lstm}. This type of network improves the simple Multilayer Perceptron (MLP) network by including an output that depends on historical learned informations. The LSTM architecture is characterized by hidden units, called memory blocks. These units allow the network to remember information over short/long sequences.
Moreover, these gates allow the LSTM model to beat the issues that inhibit the training of other deep models including RNNs and MLPs. This, and therefore the impressive results that may be achieved, are the rationale for its popularity on an outsized sort of problems \cite{choi2017using,reimers2017optimal}.

\subsection{Model Configuration}

The efficient implementation of deep learning requires the selection and optimization of many hyperparameters, as well as extensive trial and error to find the optimal values. In order to assess the advanced performance, data is divided into training, test and validation sets; The training set is being used to train learning classifier; the validation set is used to fine-tune the parameters and estimate the behavior of the classifier; and the test set is going to be used to determine the efficiency of the classifier. Once data is splitted, we tune models using K-fold cross-validation. In this study, we set K = 3. The implemented LSTM model used Tanh activation function in the hidden layers and Sigmoid activation function in the output layer. Dropout with a rate of $0.2$ is used as a regularization technique for weight optimization. In our model, a learning rate of $1e^{-03}$ is used, the number of epochs to train is set to 60 and the batch size is set to 100.

\subsection{Model Implementation}
In this work, the model was implemented using Keras framework, with TensorFlow backend. The implementation part of the proposed model consists of two stages:
\begin{enumerate}
    \item Feature Engineering: we chose big data tool like Apache Hive 2.1.0 on Microsoft Azure remote cluster (2 head nodes and 1 worker node, each with 200 GB space, 14 GB RAM, and 4 processors), to perform data preprocessing and feature engineering.
    \item Deep Learning using Colaboratory.
\end{enumerate}
We also used Python and several packages for efficient model testing, hyperparameter tuning and model evaluation including: Pandas, NumPy, SciPy, Scikit-learn, Matplotlib, Seaborn.

\section{Experimental Results}\label{exp}
In this section, we describe the results of our experiments by evaluating the LSTM model against the traditional state of the art acuity scores and machine learning approaches that were used to predict possible future clinical events supported time-series measurements, including SOFA score, SAPS-II score, SL, SVM, LR, NB and CNN. Individual sets of parameters were tuned using 3-fold cross-validation to evaluate the potency of every fixed model. Experiments were conducted under three settings: using temporal physiological measures within 6-hour, 12-hour, and 24-hour time-frames. It's worth noting that SAPS-II and SOFA acuity scores use the primary 24 hours of data to evaluate patient severity of illness.

For binary-classifier, we opt for F1-score and MCC metrics to evaluate the effectiveness of the model. In gist, these two metrics were chosen because they provide a more realistic measure of a model’s performance, and hence they are robust for binary classification problems \cite{metrics}.

Results outputs of different classifiers are presented in Table \ref{ta5}. In the light of the obtained results, fitting an LSTM model on the multivariate time-series records within a 6-hour time-frame has improved the prediction of early diagnosis of mortality risk for patients who remained in intensive departments. In fact, it is often seen from Table \ref{ta5} that the LSTM model under the tuned configuration features a higher F1-score and MCC compare to the opposite mortality predictive approaches, which approved that the performance of the LSTM model is more consistent. Although the CNN model has attained a better F1-score and MCC within a 24-hour time-frame , the LSTM model outperformed it within 6-hour and 12-hour time-frames, validating its potency in predicting mortality risk as soon as possible following the admission of patients to the critical units.
\begin{table}[tbh!]
\centering
\caption{Mortality prediction performance for binary-classification approaches (The best performing model is highlighted in \textbf{bold}).}
\begin{tabular}[t]{l | cccccc}
\multirow{3}{*}{\rotatebox{45}{\textbf{Classifier}}} &
\multicolumn{6}{c}{\textbf{Observation periods}}
\\\cline{2-7}

& \multicolumn{2}{c}{6-hour} & \multicolumn{2}{c}{12-hour} & \multicolumn{2}{c}{24-hour}\\
\cline{2-7}
  & \textbf{F1-score} & \textbf{MCC} &  \textbf{F1-score} & \textbf{MCC} &  \textbf{F1-score} & \textbf{MCC}\\ [0.5ex]
\hline\hline
SAPS-II & - & - &  - & - &  $0.41$ & $0.33$ \\\hline
SOFA    & - & -   &- & - &   $0.06$ & $0.14$ \\\hline
NB  & $0.44$ & $0.36$  & $0.55$ & $0.49$  & $0.55$ & $0.49$ \\\hline
LR  & $0.09$ & $0.20$ &  $0.14$ & $0.25$  & $0.17 $ & $0.28$ \\\hline
SVM & $0.33$ & $0.20$ & $0.30$ & $0.18$  & $0.27$ & $0.17$ \\\hline
SL  & $0.55$ & $0.57$  & $0.65$ & $0.66$  & $0.67$ & $0.68$  \\\hline
CNN & $0.92$ & $0.91$ & $\textbf{0.97}$ & $0.96$ &  $\textbf{0.99}$ & $\textbf{0.99}$ \\\hline
LSTM & $\textbf{0.96}$ & $\textbf{0.95}$ & $\textbf{0.97}$ & $\textbf{0.97}$ &  $0.96$ & $0.96$ \\
\hline
\end{tabular}
\label{ta5}
\end{table}

\begin{figure}[tbh!]
\centering
\includegraphics[width=1\columnwidth]{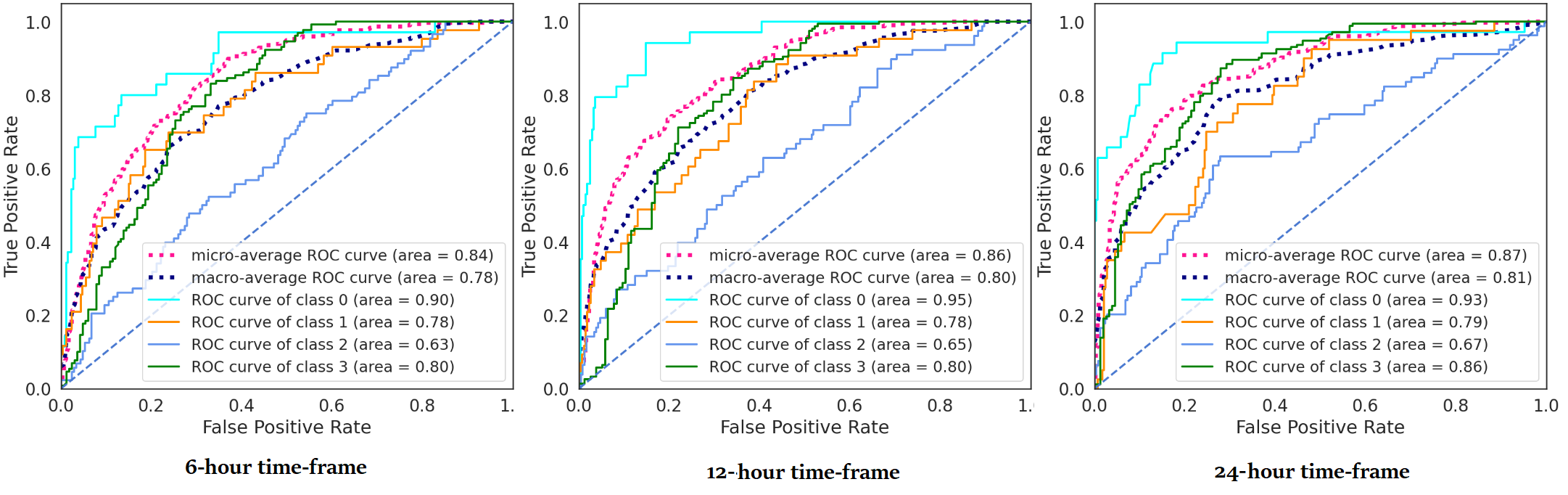}
\caption{ROC curves of the LSTM model fitted on data aggregated over 6-hour (in the left), 12-hour (in the middle), and 24-hour time-frames (in the right), applied for the multi-classification problem.}
\label{figg2}
\end{figure}

Regarding multi-class classification, the average of the evaluation measures can provides a view on the overall results for the potency of LSTM fitted on the data aggregated over 6-hour within the prediction of LOS compared to those aggregated over 12-hour and 24-hour time-frames. Two major names to refer to averaged results are micro-average and macro-average.
In gist, a macro-average will compute the metric independently for every class then take the average, whereas a micro-average will aggregate the contributions of whole classes to compute the average metric. Fig. \ref{figg2} summarizes Micro and Macro-average results for AUROC metrics and confirms that multivariate time-series data aggregated over a 6-hour time-frame offer rigorous multi-classification results compared with 12-hour and 24-hour time-frames that indicate slight improvement results.

\vspace{-6mm}

\section{Conclusion and Future Works}\label{con}
Enhancing the excellence of care for patients and predicting future outcomes are the foremost important targets in critical care research. In this paper, and by deploying multivariate time-series data obtained from EHR-database MIMIC-III, we reveal that the LSTM model systematically outperforms all opposing predictive models of mortality using physiological measures observed during 6 and 12 hours. These positive results recommend that access to the patient's physiological data trajectory as early as possible could enhance the potential in monitoring and predicting possible future events concerning the patient’s conditions in ICUs.
In future work, we arrange to apply the proposed model in other clinical tasks including early triage and risk assessment, prediction of physiologic decompensation, and identification of high-cost patients.

\section{Acknowledgement}
This work was supported by the Ministry of Higher Education and Scientific Research of Tunisia through the PEJC Young Researchers
Encouragement Program (Project code 20PEJC 05-16).
%
%
%
%

\end{document}